\pdfoutput=1

\documentclass[11pt]{article}

\usepackage[preprint]{acl}

\usepackage{times}
\usepackage{latexsym}
\usepackage{booktabs}
\usepackage{graphicx} 

\usepackage[T1]{fontenc}

\usepackage[utf8]{inputenc}

\usepackage{microtype}

\usepackage{inconsolata}

\DeclareUnicodeCharacter{2212}{-}

\newcounter{HWNumberOfComments}
\stepcounter{HWNumberOfComments}

\newcounter{JSNumberOfComments}
\stepcounter{JSNumberOfComments}

\title{\textsc{ToBlend}: Token-Level Blending With an Ensemble of LLMs to Attack AI-Generated Text Detection}

\author{Fan Huang, Haewoon Kwak, Jisun An \\
  Indiana University Bloomington / United States \\
  \texttt{huangfan@acm.org, haewoon@acm.org, jisun.an@acm.org}}

\begin{document}
\maketitle
\begin{abstract}
The robustness of AI-content detection models against sophisticated adversarial strategies, such as paraphrasing or word switching, is a rising concern in natural language generation (NLG) applications. 
This study proposes \textsc{ToBlend}, a novel token-level ensemble text generation method to challenge the robustness of current AI-content detection approaches by utilizing multiple sets of candidate generative large language models (LLMs). 
By randomly sampling token(s) from candidate LLMs sets, we find \textsc{ToBlend} significantly drops the performance of most mainstream AI-content detection methods. 
We evaluate the text quality produced under different \textsc{ToBlend} settings based on annotations from experienced human experts. 
We proposed a fine-tuned Llama3.1 model to distinguish the \textsc{ToBlend} generated text more accurately. 
Our findings underscore our proposed text generation approach's great potential in deceiving and improving detection models. Our datasets, codes, and annotations are open-sourced\footnote{\url{https://anonymous.4open.science/r/ToBlend/}}.

\end{abstract}

\section{Introduction}

The pervasiveness of generative AI has reshaped information creation and dissemination online. Powerful LLMs like ChatGPT~\cite{openai_chatgpt} and Llama~\cite{touvron2023llama} have blurred the lines between human-authored and machine-generated content~\cite{sadasivan2023can}. While these advancements bring opportunities for natural language understanding and content creation~\cite{gilardi2023chatgpt,qin2023chatgpt}, they also pose challenges~\cite{bang2023multitask}, particularly in misinformation~\cite{chi2023ai_misinfo}, copyright violation~\cite{karamolegkou-etal-2023-copyright}, and decision trustworthiness~\cite{choudhury2023investigating}. Detecting AI-generated content is thus crucial to maintaining information integrity.

Existing works on AI content detection can be broadly categorized into supervised classifiers~\cite{solaiman2019release,fagni2021tweepfake,mitrovic2023chatgpt}, adversarial learning models~\cite{radar2023,outfox_2024}, and zero-shot classifiers~\cite{gehrmann2019gltr,mitchell2023detectgpt,su2023detectllm}. These methods face constant challenges from adversarial techniques like character substitution (homoglyphs), misspelling, paraphrasing, and word-switching~\cite{wolff2020attacking,sadasivan2023can,mao2024raidar}, which have shown varying degrees of effectiveness. Recently, DetectGPT~\cite{mitchell2023detectgpt} and Fast-DetectGPT~\cite{bao2023fast} achieved outstanding detection accuracy at reduced costs, using conditional probability curvature to differentiate LLM and human word usage, and have proven resilient to mainstream adversarial attacks.

However, existing detection methods focus on detecting text generated by a single model~\cite{bao2023fast}, which may not hold in practical scenarios. In real-world applications, texts could be generated or edited by multiple models, either sequentially or collaboratively, leading to mixed linguistic features and more complex generation patterns. This limits the effectiveness of current detectors when applied to such texts, reducing their robustness and generalizability.

To address this, we propose \textsc{ToBlend}, a novel token-level ensemble text generation method, to validate the robustness of existing mainstream AI-generation detectors. \textsc{ToBlend} manipulates token selection and alters the next-token probability distribution, enabling adversaries to challenge detection accuracy effectively. Driven by the need to explore and mitigate vulnerabilities in current AI-content detection methodologies~\cite{sadasivan2023can,mitchell2023detectgpt}, we provide empirical evidence that \textsc{ToBlend} significantly degrades detection performance by exposing weaknesses in assuming a single-model generation process.

We thoroughly assess the effectiveness and limitations of \textsc{ToBlend} through four research questions: RQ1. How significantly does \textsc{ToBlend} disrupt detection models? RQ2. Do More Advanced LLMs Improve the Efficacy of \textsc{ToBlend}? RQ3. How contextually coherent and fluent is the generated text by \textsc{ToBlend}? and RQ4. Can text generated by \textsc{ToBlend} be used to improve AI-generated text detection methods?

The contributions of this paper are as follows:
First, we demonstrate empirically that \textsc{ToBlend} significantly impacts AI-generated text detection models' performance, revealing weaknesses in current detection strategies while maintaining text quality comparable to baseline models. 
Second, we show that more advanced candidate LLMs create greater disruption for detection models.
Thirdly, we report that high-quality instance pairs generated by \textsc{ToBlend} can improve AI-generated text detection performance through fine-tuning.
Our findings offer insights for future research to enhance the robustness of AI-generated text detection techniques against evolving adversarial attacks.

\begin{figure*}[th!]
    \centering
    \includegraphics[width=\textwidth]{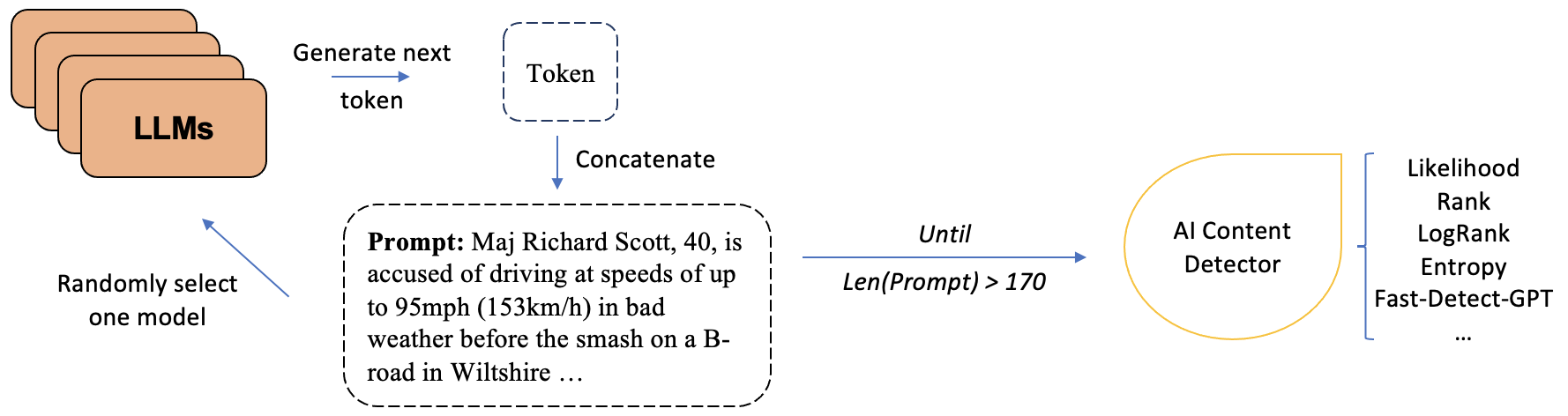}
    \caption{Pipeline illustration of \textsc{ToBlend}}
    \label{fig-pipeline}
\end{figure*}

\section{Related Work}

\noindent \textbf{AI-Generated Content and Detection Models.}
Along with the advancements in content generation~\cite{qin2023chatgpt}, efforts have been made to develop detection models capable of distinguishing human-written from AI-generated texts~\cite{mitchell2023detectgpt, bao2023fast}. Techniques leveraging word entropy analysis, machine learning classifiers, and next-token probability analysis have been explored~\cite{kirchenbauer2023watermark, tang2023science}. For instance, \citet{tang2023science} have demonstrated using statistical methods and fine-tuned models to improve detection accuracy. However, these methods often struggle against straightforward manipulation strategies, such as paraphrasing attacks, highlighting the gap in the current detection capabilities~\cite{sadasivan2023can}.

\noindent \textbf{Adversarial Attacks on AI-Content Detection.}
The concept of adversarial attacks in AI content detection involves manipulating input textual information to deceive detection models into misclassifying AI-generated content as human-written~\cite{sadasivan2023can}. 
\citet{bao2023fast} and \citet{mitchell2023detectgpt} have shed light on the vulnerabilities of AI models to adversarial inputs, suggesting that even minor alterations can significantly impact model performance. \citet{krishna2024paraphrasing} showcased the efficacy of paraphrasing attacks towards AI generation detection models and applicable retrieval-based solutions. Additional approaches like \citet{liu2022coco} and \citet{mao2024raidar} also perform well, providing feasible solutions to detect AI-generated texts effectively.

\section{\textsc{ToBlend}}

\textsc{ToBlend} is a cultivated adversarial strategy designed to deceive AI-content detection models by exploiting their reliance on predicting the next-token distribution, as illustrated in Figure~\ref{fig-pipeline}. When generating the next token given the previous text, we randomly select one LLM from a pool of multiple LLMs and let it generate the next token(s), inspired by the \textit{ensemble} method. This process repeats until the generated text meets an ending condition (e.g., its length becomes longer than the threshold). 
As a result, \textsc{ToBlend} chooses the token by shuffled probability distributions across candidate LLMs, creating a mixture of various token distribution predictions.

We carefully set the completion criterion during \textsc{ToBlend} based on previous work. In \citet{bao2023fast}, the authors controlled AI text generation by the total token length, more than 50 and less than 200 tokens, where the initial prompt includes the first 30 tokens of human-written text. Similarly, we set our completion criteria as `stop generating next token(s) when the content token length is greater than 170,' because the average token length of our three human-written text datasets is around it. We also explored another completion criterion, such as setting it to 150 instead of 170 tokens, and found that it did not significantly change the performance results, as shown in Appendix~\ref{sec:appendix-robustness}.

\section{Experiments}
\subsection{Dataset}
To evaluate \textsc{ToBlend}, we use three human-written text datasets from different domains, widely used in AI-generation detection~\cite{mitchell2023detectgpt,su2023detectllm,bao2023fast,radar2023,hao2024learning}: the XSum dataset~\cite{xsum-emnlp}, featuring news articles; the sQuAD dataset~\cite{rajpurkar2016squad}, based on Wikipedia documents; and the WritingPrompts dataset~\cite{fan-etal-2018-hierarchical_writingprompt}, containing story scripts. All datasets are open-sourced, and our use complies with their intended purposes.

To construct the AI-generated text dataset, following \citet{bao2023fast}, we adopt the same 300 instances from sQuAD and 500 from the XSum and WritingPrompts datasets used in \citet{bao2023fast}. We then prompt four LLMs (GPT-2~\cite{radford2019language}, OPT~\cite{zhang2022opt}, GPT-Neo~\cite{gpt-neo}, GPT-J~\cite{gpt-j}) with the first 30 tokens of the human-written text, limiting the generated text to 170 additional tokens. 

\subsection{Candidate LLMs for \textsc{ToBlend}}

\textsc{ToBlend} generates the next token by randomly selecting an LLM from a set of candidate LLMs and concatenating their outputs. To ensure fair comparison and demonstrate the technical potential of \textsc{ToBlend}, we use eight LLMs, divided into two sets: one replicating results from a previous study on AI-generated text detection~\cite{bao2023fast}, and the other comprising slightly larger models (up to 10 billion parameters) that are more advanced. The first set, referred to as \textit{classic} LLMs, includes GPT-2 (gpt2-xl, 1.5 billion parameters)\cite{radford2019language}, OPT (opt-2.7b)\cite{zhang2022opt}, GPT-Neo (gpt-neo-2.7B)\cite{gpt-neo}, and GPT-J (gpt-j-6B)\cite{gpt-j}. The second set, referred to as \textit{advanced} LLMs, includes Llama2 (llama-2-7b)\cite{touvron2023llama}, Phi-2 (microsoft/phi-2, 2.7 billion parameters)\cite{microsoft2023phi2}, Mistral (mistralai/Mistral-7B-v0.3)\cite{mistral2024}, and Gemma (google/gemma-7b)\cite{gemma2024google}. \looseness=-1

\subsection{Experiment Settings}
We chose the aforementioned eight open-sourced LLMs, which are between 1 and 10 billion parameters, to make \textsc{ToBlend} efficiently implemented on small GPU servers. We ran \textsc{ToBlend} using A100 80GB VRAM. Our system reached the highest average generation speed of 10 seconds per instance (around 170 tokens) in A100 for the classic/advanced LLMs sets.
The detailed inference settings are listed in Appendix~\ref{sec:appendix-inference}.

We comprehensively evaluate \textsc{ToBlend}'s performance with varying token lengths (i.e., \textit{k}, from 1 to 5 or a random value within this range) concatenated to the incomplete text in each iteration, as shown in Figure~\ref{fig-pipeline}. To ensure generation quality, we generate 3 more tokens than required and then select the needed length for concatenation (i.e., if \textit{k} tokens are needed, we generate \textit{k+3} tokens). This approach is particularly effective when \textit{k} is small. Example illustrations are provided in Appendix~\ref{sec:appendix-generation_length}. 
Additionally, we test a sentence-level setting, where each LLM generates an entire sentence instead of just a few tokens. 

We give the first 30 tokens of the human-written original sentence and let \textsc{ToBlend} complete it, following the practice in AI-generated text study~\cite{mitchell2023detectgpt, bao2023fast, radar2023}. For example, the prompt of the first instance in the XSum Dataset is:
\begin{quote}
   Maj Richard Scott, 40, is accused of driving at speeds of up to 95mph (153km/h) in bad weather before the smash on a B-road in Wiltshire
\end{quote}

\subsection{AI-Generated Text Detection}
We use multiple AI-generated text detection models to validate \textsc{ToBlend}. For the conventional statistical methods, we use likelihood (average log probabilities), rank (average token ranks arranged by descending order on probabilities), LogRank (average log value of ranks), and entropy (average token entropy of the generated content)~\cite{gehrmann2019gltr,solaiman2019release,ippolito-etal-2020-automatic}. We also include the supervised classifier GPT-2 Detector~\cite{solaiman2019release} and the adversarial-learning-based detector RADAR~\cite{radar2023}.
In addition to those methods, we adopt Fast-DetectGPT~\cite{bao2023fast}, which has achieved a higher speed and better AUROC score than DetectGPT~\cite{mitchell2023detectgpt}. 
Fast-DetectGPT remains robust even after the paraphrasing attack~\cite{sadasivan2023can}, ensuring its capability as a strong competitor for \textsc{ToBlend}~\cite{hao2024learning}.
We note that the detection approaches mentioned are all open-sourced. 

\looseness=-1 
We select the area under the receiver operating characteristic (AUROC) as the evaluation metric to illustrate the performance of AI-generation detection approaches~\cite{bao2023fast}. 
The effectiveness of \textsc{ToBlend} against the detection models is quantified by comparing the AUROC score of detecting text generated by previously mentioned LLMs with detecting text generated by \textsc{ToBlend}. 

\subsection{Generated Text  Quality Evaluation}
The generated text by \textsc{ToBlend} should have an equivalent quality compared with that by a single LLM. 
We randomly selected five instances for each dataset to construct a comprehensive analysis of the generation result of our proposed \textsc{ToBlend} generation approach. We then hired three well-trained human experts (one male and two female from the institution) to annotate the quality of \textsc{ToBlend} generation results in seven experimental settings and the baseline generated by the GPT-2 (i.e., \textit{gpt-xl}) and ChatGPT-3.5 (i.e., \textit{gpt-3.5-turbo-0125}).  
We use two metrics: coherence~\cite{cervone2018coherence, ye-etal-2021-towards-quantifiable} and fluency~\cite{martindale2018fluency, kann-etal-2018-sentence}, which have long been perceived as key metrics to evaluate generation results in the field of dialogue system and NLG evaluations.

\section{RQ1: How Significantly Does \textsc{ToBlend} Disrupt Detection Models?}
\label{RQ1}
\begin{table}[ht]
    \centering
    \begin{tabular}{p{15mm}|p{10.5mm}|p{10.5mm}|p{10mm}|p{10mm}}
    \toprule
    Datasets & GPT-2 & OPT & Neo & GPT-J \\
    \midrule
    XSum & 0.9922  & 0.9806 & 0.9881 & 0.9771 \\
    \hline
    sQuAD & 0.9990 & 0.9949 & 0.9956 & 0.9854 \\
    \hline
    Writing & 0.9982 & 0.9972 & 0.9981 & 0.9974 \\
    \hline
    \textbf{Avg.} & 0.9965 & 0.9909 & 0.9939 & 0.9866 \\
    \midrule
    \citet{bao2023fast} & \textit{0.9967} & \textit{0.9908} & \textit{0.9940} & \textit{0.9866}\\
    \bottomrule
    \end{tabular}
    \caption{Replication of the detection AUROC scores of Fast-DetectGPT on single LLM generation with the same settings as ~\citet{bao2023fast}. The final row lists the original results from ~\citet{bao2023fast}}
    \label{tab-replication}
\end{table}

\begin{table*}[ht]
    \centering
    \begin{tabular}{p{12mm}|p{27.5mm}|p{12mm}|p{9mm}|p{9mm}|p{9mm}|p{9mm}|p{9mm}|p{9mm}|p{9mm}}
    \toprule
    Datasets & Detection Method & Baseline & TL=1 & TL=2 & TL=3 & TL=4 & TL=5 & Rand. & Sent. \\
    \hline
    \hline
        & Likelihood & 0.7837 & \textbf{0.3444} & 0.3526 & 0.3696 & 0.3716 & 0.3738 & 0.3474 & 0.4492 \\
        & Rank & 0.8068 & \textbf{0.4845} & 0.4984 & 0.4956 & 0.5027 & 0.5066 & 0.5130 & 0.5797\\
        & LogRank & 0.8117 & 0.4201 & \textbf{0.4189} & 0.4355 & 0.4380 & 0.4339 & 0.4190 & 0.5191 \\
    XSum  & Entropy & 0.5300 & 0.7971 & 0.7803 & 0.7714 & 0.7711 & 0.7644 & 0.7792 & 0.7435 \\
        & GPT-2 Detector & 0.9416 & 0.7211 & 0.6924 & 0.7079 & 0.6922 & 0.6948 & \textbf{0.6776} & 0.7025 \\
        & RADAR & 0.8904 & 0.8958 & 0.9225 & 0.9080 & 0.9130 & 0.9142 & 0.8937 & 0.8963 \\
        & Fast-DetectGPT & 0.9845 & 0.7088 & \textbf{0.7004} & 0.7172 & 0.7291 & 0.7250 & 0.7026 & 0.8062 \\
    \hline
    \hline
        & Likelihood & 0.7573 & \textbf{0.4350} & 0.4707 & 0.4961 & 0.4847 & 0.5124 & 0.4624 & 0.5493 \\
        & Rank & 0.7836 & \textbf{0.5485} & 0.5758 & 0.5728 & 0.5801 & 0.5790 & 0.5672 & 0.6224 \\
        & LogRank & 0.8090 & \textbf{0.5281} & 0.5609 & 0.5850 & 0.5711 & 0.5993 & 0.5508 & 0.6357 \\
    sQuAD    & Entropy & 0.5617 & 0.7469 & 0.7551 & 0.7212 & 0.7333 & 0.7208 & 0.7441 & 0.7152 \\
        & GPT-2 Detector & 0.9335 & 0.7208 & 0.7027 & \textbf{0.6922} & 0.6963 & 0.7085 & 0.6991 & 0.7107 \\
        & RADAR & 0.8263 & 0.8395 & 0.8641 & 0.8670 & 0.8812 & 0.8532 & 0.8564 & \textbf{0.8067} \\
        & Fast-DetectGPT & 0.9937 & \textbf{0.8047} & 0.8643 & 0.8581 & 0.8557 & 0.8685 & 0.8282 & 0.9062 \\
    \hline
    \hline
        & Likelihood & 0.8905 & \textbf{0.7029} & 0.7293 & 0.7176 & 0.7444 & 0.7407 & 0.7123 & 0.7866 \\
        & Rank & 0.8186 & \textbf{0.6558} & 0.6672 & 0.6822 & 0.6826 & 0.6867 & 0.6762 & 0.7145 \\
        & LogRank & 0.9158 & \textbf{0.7585} & 0.7823 & 0.7727 & 0.7949 & 0.7904 & 0.7669 & 0.8305 \\
    Writing    & Entropy & 0.3752 & 0.5708 & 0.5795 & 0.5840 & 0.5573 & 0.5706 & 0.5885 & 0.5135 \\
        & GPT-2 Detector & 0.9365 & 0.7434 & 0.7316 & 0.7474 & 0.7405 & \textbf{0.7303} & 0.7444 & 0.7437 \\
        & RADAR & 0.7855 & 0.8846 & 0.8347 & 0.8305 & 0.8275 & 0.8203 & 0.8345 & \textbf{0.7761} \\
        & Fast-DetectGPT & 0.9977 & \textbf{0.9128} & 0.9451 & 0.9375 & 0.9439 & 0.9480 & 0.9379 & 0.9581 \\
        \hline
    \bottomrule
    \end{tabular}
    \caption{Result of the classic LLMs (GPT-2, OPT, GPT-Neo, and GPT-J)---AUROC scores for the XSum, sQuAD, and Writing datasets across various token lengths. Baseline scores come from the average score for each dataset in Table~\ref{tab-replication}. Compared with the baseline AUROC score at each row, we highlighted the most deviated AUROC score in bold. TL is token length.  Rand means that the token number is randomly chosen between 1 and 5 when generation. Sent is the sentence-level blending.}
    \label{tab-benchmark-all-metrics}
\end{table*}

We build the baseline by replicating Fast-DetectGPT's performance in detecting AI-generated content. 
We use each of the four LLMs (GPT-2, OPT, GPT-Neo, and GPT-J) to generate the AI content and use Fast-DetectGPT to detect them using the same settings in \cite{bao2023fast}. 
We successfully replicate Fast-DetectGPT's AUROC scores in our settings as shown in Table~\ref{tab-replication}. 
For each dataset, we use the average AUROC scores across the four models as a baseline when applying the classic LLMs  (Table~\ref{tab-benchmark-all-metrics}). 

\looseness=-1  
Table~\ref{tab-benchmark-all-metrics} shows that \textsc{ToBlend} notably degrades the performance of nearly all traditional AI content detection methods (such as likelihood, rank, and LogRank) and the SoTA detection models, GPT-2 Detector and Fast-DetectGPT. 
The significant drops in performance across different detection approaches highlight the effectiveness of the \textsc{ToBlend} in exploiting the inherent weaknesses of current mainstream detection techniques.
The GPT-2 Detector and Fast-DetectGPT have high AUROC scores for the baseline but show significant performance drops for the text generated by \textsc{ToBlend}. 
When using the Fast-DetectGPT detection method, the one or two token length settings perform the best, decreasing the AUROC score from 0.9845 to 0.7004 for XSum, from 0.9937 to 0.8047 for sQuAD, and from 0.9977 to 0.9128 for WritingPrompts. 
As for the GPT-2 Detector, the scores drop significantly from 0.9416 to 0.6776 for XSum, from 0.9335 to 0.6922 for sQuAD, and from 0.9365 to 0.7303 for WritingPrompts. 
Generally, the degradation of detection models is greater when \textsc{ToBlend} generates fewer tokens at each step and concatenates them.

On the other hand, the entropy and RADAR detection methods, perform better when identifying texts generated through \textsc{ToBlend} methods.
This suggests that the current ensemble of LLMs used in \textsc{ToBlend} may not reproduce the entropy distribution of the genuine text and appears ineffective at fooling detection models trained against multi-LLM paraphrasing attacks.  
We report three more AI-generated detection approaches from DetectGPT~\cite{pmlr-v202-mitchell23a} and DetectLLM (LLR and NPR methods)~\cite{su2023detectllm} in the Appendix~\ref{sec:appendix-more-detection-models}\footnote{The implementations of DetectGPT and DetectLLM models in our local server are relatively time-consuming. So we report the AUROC scores calculated on the sub-sampled dataset to improve the time efficiency}.

\looseness=-1 

\section{RQ2: Do More Advanced LLMs Improve the Efficacy of \textsc{ToBlend}?}
\label{RQ2}

\begin{table*}[ht]
    \centering
    \begin{tabular}{p{12mm}|p{27.5mm}|p{12mm}|p{9mm}|p{9mm}|p{9mm}|p{9mm}|p{9mm}|p{9mm}|p{9mm}}
    \toprule
    Datasets & Detection Method & Baseline & TL=1 & TL=2 & TL=3 & TL=4 & TL=5 & Rand. & Sent. \\
    \hline
    \hline
        & Likelihood & 0.7837 & \textbf{0.1938} & 0.2190 & 0.2036 & 0.2493 & 0.2396 & 0.2122 & 0.3023 \\
        & Rank & 0.8068 & \textbf{0.3538} & 0.4036 & 0.3723 & 0.4246 & 0.3896 & 0.3794 & 0.4642\\
        & LogRank & 0.8117 & \textbf{0.2426} & 0.2606 & 0.2498 & 0.2917 & 0.2878 & 0.2506 & 0.3389 \\
    XSum    & Entropy & 0.5300 & 0.8121 & 0.8002 & 0.8057 & 0.7723 & 0.7774 & 0.8025 & 0.7495 \\
        & GPT-2 Detector & 0.9416 & 0.5622 & \textbf{0.5411} & 0.5723 & 0.5757 & 0.5546 & 0.5414 & 0.5520 \\
        & RADAR & 0.8904 & 0.9027 & 0.9412 & 0.9423 & 0.9391 & 0.9419 & 0.9307 & 0.9075 \\
        & Fast-DetectGPT & 0.9845 & \textbf{0.3968} & 0.4479 & 0.4203 & 0.4580 & 0.4499 & 0.4285 & 0.5382 \\
    \hline
    \hline
        & Likelihood & 0.7573 & \textbf{0.2968} & 0.3451 & 0.3638 & 0.3844 & 0.3762 & 0.3541 & 0.4800 \\
        & Rank & 0.7836 & \textbf{0.4016} & 0.4297 & 0.4344 & 0.4730 & 0.4627 & 0.4472 & 0.5286 \\
        & LogRank & 0.8090 & \textbf{0.3600} & 0.4041 & 0.4267 & 0.4545 & 0.4422 & 0.4140 & 0.5360 \\
    sQuAD    & Entropy & {0.5617} & 0.7363 & 0.7096 & 0.6990 & 0.6961 & 0.6968 & 0.6974 & 0.6290 \\
        & GPT-2 Detector & 0.9335 & 0.5221 & 0.5097 & \textbf{0.5052} & 0.5367 & 0.5131 & 0.5056 & 0.5368 \\
        & RADAR & {0.8263} & 0.8474 & 0.8820 & 0.8785 & 0.8846 & 0.8862 & 0.8675 & 0.8380 \\
        & Fast-DetectGPT & 0.9937 & \textbf{0.5317} & 0.5803 & 0.5816 & 0.6178 & 0.6024 & 0.5684 & 0.7046 \\
    \hline
    \hline
        & Likelihood & 0.8905 & \textbf{0.5845} & 0.6354 & 0.6461 & 0.6543 & 0.6520 & 0.6362 & 0.7126 \\
        & Rank & 0.8186 & \textbf{0.5952} & 0.6101 & 0.6220 & 0.6358 & 0.6348 & 0.6197 & 0.6641 \\
        & LogRank & 0.9158 & \textbf{0.6358} & 0.6807 & 0.6903 & 0.6980 & 0.6966 & 0.6828 & 0.7487 \\
    Writing    & Entropy & {0.3752} & 0.6061 & 0.5619 & 0.5575 & 0.5626 & 0.5532 & 0.5766 & 0.5102 \\
        & GPT-2 Detector & 0.9365 & 0.6486 & 0.6436 & 0.6280 & 0.6303 & 0.6381 & 0.6473 & \textbf{0.6230} \\
        & RADAR & {0.7855} & 0.8844 & 0.8294 & 0.8233 & 0.8261 & 0.8407 & 0.8419 & 0.8056 \\
        & Fast-DetectGPT & 0.9977 & \textbf{0.8036} & 0.8272 & 0.8398 & 0.8572 & 0.8448 & 0.8348 & 0.8846 \\
        \hline
    \bottomrule
    \end{tabular}
    \caption{Result of the advanced LLMs set---AUROC scores for the XSum, sQuAD, and Writing datasets by various \textsc{ToBlend} generation settings. The difference is that the candidate models used here are Llama2, Phi-2, Mistral, and Gemma.}
    \label{tab-benchmark-better-all-metrics}
\end{table*}

Through replication of the AI-generated text detection study~\cite{bao2023fast} in the previous section, we showed that \textsc{ToBlend} works well even with early LLMs, such as GPT-2 and OPT. This raises a natural question:  How much can performance be improved when \textsc{ToBlend} is used with more advanced LLMs? 
The answer to this question can also illustrate \textsc{ToBlend}'s potential as newer LLMs are released in the future.

The experiment results using the advanced LLMs are presented in Table~\ref{tab-benchmark-better-all-metrics}. 
Comparing the AUROC scores among various settings and datasets from Table~\ref{tab-benchmark-all-metrics} and Table~\ref{tab-benchmark-better-all-metrics}, as one can expect, \textsc{ToBlend} with the advanced LLMs is more successful at deceiving the AI-generation detection approaches when it was effective. 
When testing on the Fast-DetectGPT method, the one- or two-\textsc{ToBlend} setting performs even better than in Table~\ref{tab-benchmark-all-metrics}, decreasing the AUROC score from 0.9845 to 0.3968 for XSum, from 0.9937 to 0.5317 for sQuAD, and from 0.9977 to 0.8036 for WritingPrompts. 
As for the GPT-2 Detector, the scores drop significantly from 0.9416 to 0.5411 for XSum, from 0.9335 to 0.5052 for sQuAD, and from 0.9365 to 0.6230 for WritingPrompts. 

However, the two methods that were not affected by \textsc{ToBlend}, Entropy and RADAR, still show similar patterns with advanced LLMs.

The variation in performance is likely due to the different syntactic and semantic patterns these models introduce, as showcased in Appendix~\ref{sec:appendix-filter-function}. More significant AUROC score drops in Table~\ref{tab-benchmark-better-all-metrics} further support the assumption that the more advanced the models in the blending, the lower the likelihood of accurate detection, especially for the current SOTA detection model like Fast-DetectGPT.
We report a few more detection performance of DetectGPT~\cite{pmlr-v202-mitchell23a} and DetectLLM~\cite{su2023detectllm} in the Appendix~\ref{sec:appendix-more-detection-models}.

\section{RQ3: How Contextually Coherent and Fluent Are the Text Generated by \textsc{ToBlend}?}
\label{RQ3}

\begin{table*}[ht]
    \centering
    \begin{tabular}{p{11.5mm}|p{16.5mm}|p{10.5mm}|p{13mm}|p{9mm}|p{9mm}|p{9mm}|p{9mm}|p{9mm}|p{9mm}|p{8.5mm}}
    \toprule
    Datasets & Candidate LLMs & GPT-2 & ChatGPT & TL=1 & TL=2 & TL=3 & TL=4 & TL=5 & Rand. & Sent. \\
    \hline
    \hline
    XSum    & Classic & 5.1/5.0 & 6.3/6.7 & 4.3/4.3 & 3.9/4.3 & 3.9/4.1 & 2.8/2.9 & 5.1/5.5 & 5.7/6.0 & 4.7/4.9 \\
        & Advanced & 5.1/5.0 & 6.3/6.7 & 4.1/4.4 & 4.1/4.6 & 4.2/4.5 & 4.1/4.3 & 4.8/5.1 & 5.1/6.1 & 4.2/4.3 \\
    \hline
    \hline
    sQuAD    & Classic & 5.4/5.4 & 6.1/6.5 & 4.3/4.4 & 4.1/4.3 & 4.1/4.2 & 4.7/4.7 & 3.9/4.3 & 5.3/5.7 & 4.1/4.9 \\
        & Advanced & 5.4/5.4 & 6.1/6.5 & 4.6/4.8 & 4.7/4.9 & 4.1/4.4 & 4.3/4.7 & 4.6/4.9 & 5.1/5.2 & 4.8/4.9 \\
    \hline
    \hline
    Writing    & Classic & 3.8/4.2 & 5.7/6.1 & 4.5/4.7 & 3.9/4.1 & 4.3/4.7 & 4.1/4.4 & 4.6/4.9 & 5.4/5.3 & 4.6/4.9 \\
        & Advanced & 3.8/4.2 & 5.7/6.1 & 4.3/4.5 & 4.5/4.6 & 4.9/5.0 & 4.3/5.0 & 4.2/4.5 & 4.7/4.9 & 4.3/4.7 \\
        \hline
    \bottomrule
    \end{tabular}
    \caption{The average coherence/fluency scores for sampled XSum, sQuAD, and Writing sub-datasets, annotated by three hired human experts and recorded in various settings of \textsc{ToBlend}.}
    \label{tab-human-evaluation}
\end{table*}

\begin{table*}[ht]
    \centering
    \begin{tabular}{p{12mm}|p{27.5mm}|p{12mm}|p{9mm}|p{9mm}|p{9mm}|p{9mm}|p{9mm}|p{9mm}|p{9mm}}
    \toprule
    Datasets & Fine-tune Status& Baseline & TL=1 & TL=2 & TL=3 & TL=4 & TL=5 & Rand. & Sent. \\
    \hline
    \hline
    XSum    & w/o fine-tuned & 0.50 & 0.53 & \textbf{0.56} & 0.52 & 0.49 & 0.53 & 0.50 & 0.45\\
        & w/ fine-tuning & \textbf{0.57} & \textbf{0.58} & 0.55 & \textbf{0.54} & \textbf{0.57} & \textbf{0.57} & \textbf{0.57} & \textbf{0.49} \\
    \hline
    \hline
    sQuAD    & w/o fine-tuned & 0.45 & \textbf{0.57} & 0.57 & \textbf{0.57} & 0.50 & \textbf{0.54} & \textbf{0.61} & 0.34 \\
        & w/ fine-tuning & \textbf{0.49} & 0.54 & \textbf{0.60} & 0.51 & \textbf{0.57} & 0.51 & 0.52 & \textbf{0.40} \\
    \hline
    \hline
    Writing    & w/o fine-tuned & 0.44 & \textbf{0.54} & \textbf{0.57} & \textbf{0.57} & 0.51 & \textbf{0.53} & \textbf{0.54} & 0.37 \\
        & w/ fine-tuning & \textbf{0.58} & 0.47 & \textbf{0.57} & 0.55 & \textbf{0.56} & 0.49 & 0.52 & \textbf{0.48} \\
        \hline
    \bottomrule
    \end{tabular}
    \caption{The accuracy of utilizing the Llama3.1 (llama-3.1-8B-Instruct) for the AI-generation detection task on the sampled XSum, sQuAD, and Writing sub-datasets before and after the specific fine-tuning process based on the selected high-quality machine-generated instances.}
    \label{tab-llm-improvement}
\end{table*}

One essential aspect of deploying adversarial strategies like \textsc{ToBlend} is ensuring that the generated text not only deceives detection systems but also retains good coherence and contextual fluency comparable to the level of human-written text. 
This section examines the text quality produced by \textsc{ToBlend} to assess its capability and limitations. 

We employed two linguistic quality metrics, which are \textit{coherence} and \textit{fluency}, from human annotations to evaluate the quality of the generated text and its resemblance to the human-written level. 
The coherence measure evaluates whether the contextual information of a given short text makes logical sense, and the fluency measure tests whether it reads naturally, mimicking the style and syntax of human language.
We collected annotations from three human experts to evaluate the quality of the text generated by \textsc{ToBlend} on a scale of 1 to 7 (1 means poor, 7 means excellent quality).
We randomly select five instances for each dataset and report the average score of five instances in different generation settings, as listed in Table~\ref{tab-human-evaluation}.

\looseness=-1  Among the baseline generation from the GPT-2 and ChatGPT-3.5 models and \textsc{ToBlend} generation in seven settings for each instance, we then asked the human experts to select the one they think has the highest probability of being written by a real human. Note that we disclose neither the source of the given short text nor our experimental settings before annotation. 
The annotation process took 14 working hours, and we reimbursed 280 US dollars. The detailed settings of human annotation collection and our instruction to the human experts are listed in Appendix~\ref{sec:appendix-human_experts}.

We can observe a high relevance between the coherence and fluency metric scores; the fluency score is mostly higher than the coherence score, indicating \textsc{ToBlend} generates more fluent but less coherent text. 
While the baseline approach receives high coherence and fluency scores, \textsc{ToBlend} remains of relatively good quality, especially when the token length increases. 
It is worth noting that utilizing advanced LLMs does not improve the quality of the generated text, even though \textsc{ToBlend} with advanced LLMs makes the generated text more indistinguishable from human-written text as shown in Table~\ref{tab-benchmark-all-metrics} and Table~\ref{tab-benchmark-better-all-metrics}.
While some token length settings produced text with high linguistic quality scores (e.g., sentence-level and random token length blending), others displayed acceptable scores in fluency and coherence, except when adopting the token length of 3 when utilizing the classic LLMs for the XSum dataset.
We attach the examples of the generated text selected to be annotated in Appendix~\ref{sec:appendix-human_experts}.

\section{RQ4: Can Text Generated by \textsc{ToBlend} Be Used to Improve AI-generated Text Detection Methods?}
\label{RQ4}
A critical challenge in AI-generated text detection is improving the accuracy and robustness of the detection model. In this section, we investigate whether fine-tuning the LLMs using high-quality instances generated by \textsc{ToBlend} can improve their ability to detect the text generated by \textsc{ToBlend}.

We implemented and fine-tuned the Llama3.1 model with 8 billion parameters (llama-3.1-8b-instruct), utilizing the LoRA~\cite{hu2021lora} for higher speed and lower GPU memory consumption. 
We filter the high-quality generation results from the 240 instances based on the quality annotation collected in Section~\ref{RQ3}. Only the instances with coherence and fluency scores equal to or higher than 5 are selected for the fine-tuning dataset. 
The detailed prompt and fine-tuning settings are attached in Appendix~\ref{sec:appendix-llama2}.

The dataset we used to fine-tune the llama3.1 model consists of 37 high-quality instances (28 from sQuAD, 7 from XSum, and 2 from Writing) generated by \textsc{ToBlend}, and another 37 human written instances from the original XSum, sQuAD, and Writing datasets (28 from sQuAD, 7 from XSum, and 2 from Writing). We then select 100 instances from each of the three datasets to form the test set to calculate the detection accuracy of the binary classification. The results in Table~\ref{tab-llm-improvement} showcase the accuracy improvements in the majority of the settings. The baseline score is the average accuracy when the text is completed by a single model in the classic LLMs.

Under the baseline settings here, we observe a significant improvement of the Llama3.1 model in detecting the AI-generated text after fine-tuning. The improvement is not stable for detecting the text generated by \textsc{ToBlend}. 

The improved accuracy in baseline and some configurations of \textsc{ToBlend} indicate the promising benefits through fine-tuning the state-of-the-art LLMs (i.e., llama3.1), which can effectively learn the high-quality AI-generated text by \textsc{ToBlend}. 

\section{Discussion}

Our research introduces \textsc{ToBlend}, a novel token-level blending text generation method, which challenges the fundamental assumption of current major detection models---that AI-generated text originates from a single model. The results demonstrate that our approach effectively disrupts the detection performance of state-of-the-art models, particularly the GPT-2 Detector and Fast-DetectGPT, by producing text with blended features that confound single-source detection methods. The increased ability of \textsc{ToBlend} to evade detection when using more advanced LLMs than classic LLMs also indicates the potential of \textsc{ToBlend} with future LLMs. Because \textsc{ToBlend} leverages an ensemble of LLMs without requiring additional training or datasets, it is readily adaptable to and can benefit from future, more advanced LLMs.

Interestingly, our approach showed limited effectiveness against two methods: Entropy and RADAR. The Entropy method, which measures whether the probability distribution is as expected, performed better in detecting our generated texts. This suggests that \textsc{ToBlend} may be producing more (or less) predictable text patterns, inadvertently making them more detectable by entropy-based methods. RADAR, on the other hand, employs a GAN-like structure with two players: a detector and a paraphraser trained on datasets constructed by paraphrasing human-written texts using various models. This approach appears to be inherently more robust against multi-model generated text, aligning closely with our method's objectives. However, RADAR's high computational cost for training limits its widespread adoption.

Despite these limitations, we believe our approach remains significant. Our findings highlight the importance of incorporating diverse detection models, including traditional entropy-based and adversarial learning-based methods, to create more robust detection systems. This diversity in detection approaches can help mitigate the risk of attacks that leverage ensembled results from multiple LLMs.

Implementing \textsc{ToBlend} reveals several trade-offs, particularly regarding generation speed, resource consumption, and the balance between deception capability and text quality. These trade-offs highlight the complexities involved in designing effective and efficient adversarial AI strategies. 

Another critical trade-off emerges between the strength of deception and the coherence and fluency of the generated text. Certain configurations of \textsc{ToBlend} (e.g., TL=1) proved highly effective in deceiving state-of-the-art LLM-based detection methods like Fast-DetectGPT. However, these configurations often resulted in less coherent and fluent text. A potential solution lies in finding a balance where the deception is successful (e.g., achieving a random guess success rate) while maintaining text quality comparable to other LLM outputs. \looseness=-1

Our research also revealed an unexpected benefit: the high-quality instance pairs generated by \textsc{ToBlend} can be used to improve AI-generated text detection performance through fine-tuning. This finding suggests that our method, while designed as an adversarial technique, may also contribute to enhancing detection capabilities.

\section{Conclusion}

This study introduces \textsc{ToBlend}, a novel attack strategy for AI content detection using a token-level blending approach. By leveraging multiple LLMs, we effectively challenge current detection models. Inspired by ensemble methods in machine learning, our approach manipulates candidate selection from a diverse array of LLMs. 

Our findings explore a new paradigm for text generation and offer insights to improve current AI-generation detection models. Key observations include:
1) \textsc{ToBlend} generated texts are significantly harder to detect while maintaining coherence and fluency;
2) Different LLM candidates produce varied results, with advanced models not always ensuring better text quality; and 
3) Simple fine-tuning improves detection models' ability to identify nuances introduced by \textsc{ToBlend}. 

These results highlight potential vulnerabilities in current detection models and emphasize the need for more robust, diverse strategies. As AI text generation evolves, detection methods must adapt to increasingly sophisticated techniques, including those using multiple models. Future research should focus on developing detection systems that can identify text generated or edited by multiple models, ensuring the integrity of online information in an era of advanced AI.

\clearpage

\section*{Ethical Statements}
The author's Institutional Review Board has approved the experiment design of collecting human annotations, and the approval number will be disclosed in the camera-ready version. We provide all annotators with information on mental consultant hotlines and clinics, considering that the LLMs generation results might contain uncomfortable information like social bias. We suggest the annotators stop or quit the annotation process anytime if they feel necessary. 
The annotators are reimbursed based on their recorded working hours at a rate above the average salary requirement in the US. 

Regarding the ethical concerns associated with AI content generation and detection, addressing the various dimensions of risk, fairness, privacy, and security issues is imperative. We want to outline the potential ethical considerations of our work, underscoring the drawbacks of misuse and possible negative consequences.

Our research, primarily technical explorations, opens trails to potentially harmful applications. 
The \textsc{ToBlend} text generation method, which was practical and straightforward to deploy, could be easily adopted to deceive the current AI content detection services, which would raise concerns regarding the spread of disinformation or the creation of fake user profiles. 
Such risks highlight the importance of developing robust detection mechanisms to identify and mitigate adversarial attacks. Mitigation strategies might include the development of more sophisticated detection algorithms, implementing ethical guidelines for AI-generated content, or promoting transparency in AI deployments. 

Regarding fairness, deploying technologies that leverage LLMs' deception capability could inadvertently amplify the misuse of LLMs on their inherent biases toward historically marginalized groups or minority groups. 
Our research methodology and applications should be carefully scrutinized to avoid bias issues, ensuring that the development and deployment of generative AI models in our experimental settings do not exacerbate social inequalities.
Exploring adversarial attacks in AI-content detection applications could also involve privacy and security considerations. The inadequate practices could inadvertently facilitate malicious activities without scrutinizing the content.

\section*{Limitations}
The findings of our study highlight a significant vulnerability in current AI-content detection models when faced with sophisticated adversarial attack strategies. Our proposed method has proven effective in degrading the performance of SOTA detection approaches. The ability of \textsc{ToBlend} generated texts to deceive detection underscores the complexity of distinguishing between human and AI-generated content, which the evolving capabilities of generative AI technologies would magnify.
As AI technologies advance, the potential for misuse in spreading misinformation or reinforcing social bias through generated deceptive content could keep increasing. Though the special fine-tuned LLM could perform better towards specific tasks and datasets, that pipeline may not work for other scenarios. Further investigations on the efficacy and robustness of fine-tuning LLM are expected. 

While our study provides valuable insights, it is important to acknowledge the limitations of our work. 
The scope of our experiments was constrained by the selection of finite LLM candidates and finite detection methods from a wide range of options. We selected the most mainstream LLMs and detection methods, while a more comprehensive benchmark would be better to illustrate the capability and limitations of our proposed \textsc{ToBlend} generation approach.
Moreover, the datasets used in our experiments may not fully capture the diversity of human and AI-generated content encountered in real-world scenarios. Including more varied and nuanced datasets could improve the analysis of the effectiveness of our proposed \textsc{ToBlend} attack. Finally, exploring alternative adversarial strategies and their countermeasures can provide a broader perspective on the race between AI content generation and detection.

\bibliography{reference.bib}

\appendix
\clearpage

\begin{table*}[ht]
    \centering
    \begin{tabular}{p{12mm}|p{27.5mm}|p{12mm}|p{9mm}|p{9mm}|p{9mm}|p{9mm}|p{9mm}|p{9mm}|p{9mm}}
    \toprule
    Datasets & Detection Method & Baseline & TL=1 & TL=2 & TL=3 & TL=4 & TL=5 & Rand. & Sent. \\
    \hline
    \hline
        & Likelihood & 0.7837 & 0.2176 & 0.1543 & 0.1988 & 0.2383 & 0.2537 & \textbf{0.1883} & 0.4087 \\
        & Rank & 0.8068 & \textbf{0.3073} & 0.3548 & 0.3970 & 0.4108 & 0.4144 & 0.3844 & 0.5610\\
    XSum & LogRank & 0.8117 & 0.2719 & \textbf{0.2254} & 0.2811 & 0.3221 & 0.3314 & 0.2574 & 0.4725\\
        & Entropy & 0.5300 & 0.7623 & 0.8467 & 0.8602 & 0.8438 & 0.8258 & 0.8489 & 0.7523 \\
        & Fast-DetectGPT & 0.9845 & \textbf{0.3187} & 0.4053 & 0.5363 & 0.5833 & 0.5897 & 0.4971 & 0.7559 \\
    \hline
    \hline
        & Likelihood & 0.7573 & \textbf{0.1812} & 0.2722 & 0.2949 & 0.3208 & 0.3775 & 0.2684 & 0.5305 \\
        & Rank & 0.7836 & \textbf{0.3117} & 0.3825 & 0.4294 & 0.4648 & 0.4908 & 0.4089 & 0.5975 \\
    sQuAD & LogRank & 0.8090 & \textbf{0.2523} & 0.3341 & 0.4101 & 0.4294 & 0.4906 & 0.3829 & 0.6053 \\
        & Entropy & 0.5617 & 0.8262 & 0.8306 & 0.7945 & 0.8017 & 0.7707 & 0.8107 & 0.7181 \\
        & Fast-DetectGPT & 0.9937 & \textbf{0.3986} & 0.5583 & 0.6401 & 0.6905 & 0.7433 & 0.6104 & 0.8974\\
    \hline
    \hline
        & Likelihood & 0.8905 & \textbf{0.5637} & 0.6017 & 0.6315 & 0.6612 & 0.6786 & 0.6287 & 0.7361 \\
        & Rank & 0.8186 & \textbf{0.5737} & 0.6140 & 0.6328 & 0.6389 & 0.6449 & 0.6402 & 0.6873 \\
    Writing & LogRank & 0.9158 & \textbf{0.6263} & 0.6748 & 0.7065 & 0.7269 & 0.7418 & 0.7030 & 0.7843\\
        & Entropy & 0.3752 & 0.5694 & 0.6177 & 0.6477 & 0.6301 & 0.6118 & 0.6221 & 0.5740 \\
        & Fast-DetectGPT & 0.9977 & \textbf{0.7168} & 0.8378 & 0.8965 & 0.9105 & 0.9106 & 0.8680 & 0.9430\\
        \hline
    \bottomrule
    \end{tabular}
    \caption{All AI content detection metric AUROC scores for the XSum, sQuAD, and Writing datasets, reported in various \textsc{ToBlend} generation settings. Baseline scores come from the average score for each dataset in Table~\ref{tab-replication}. Here we adopt the different \textsc{ToBlend} generation completion criteria of not exceeding 100 tokens ended by a period or exceeding 150 tokens for each instance. Candidate LLMs are from the classic LLMs set.}
    \label{tab-benchmark-all-metrics-no_length}
\end{table*}

\section{Inference Settings and Efficiency Analysis}
\label{sec:appendix-inference}

Our \textsc{ToBlend} generation approach does not necessarily require GPU resources. As we tested on the CPU server, the one-token \textsc{ToBlend} generation setting would need around 30 minutes to generate 100 tokens without specific speed optimization. However, using the A100 GPU server to accelerate the generation speed would only take approximately 10 seconds to generate 170 tokens in all \textsc{ToBlend} settings, which would take up to 100GB of the GPU memory usage of two A100 GPU 80GB GPU cards. We completed the experiments using the GPU server provided by Nvidia, under the support of the grant [anonymized]. (Detailed grant information will be revealed in the camera-ready version.)

\section{Token Length Settings with Examples}
\label{sec:appendix-generation_length}
In our original settings, we asked the candidate LLM to generate exactly the amount of tokens we required, i.e., if we set the token length is 1, we simply ask the model to generate only one token in each iteration. The results are listed in Table~\ref{tab-old-benchmark-all-metrics} and Table~\ref{tab-old-benchmark-better-all-metrics}. Though we have more significant AUROC score drops here, we find the generated text is of relatively lower quality compared with GPT-2 generation results, as listed in Table~\ref{tab-old-human-evaluation}. For example, below is one generation example without the extra buffer tokens (sQuAD dataset, classic LLMs set, token length for 1, without buffer tokens):

\begin{quote}
   The success of its football team made Notre Dame a household name . The success of Note Dame reflected rising status of Irish Americans and Catholics in the 1920s . Over the previous decade, Notre Dame has been the main player in the university athletic scene, and in 1922 the University took over all the sports as Notre Dame was moving up the ladder to the highest level of college athletics. < /s >> Source:< http : // www. j j r a b e r r e d o n. u a e n d o n a t s. f o r a c y. o r u s t o c h. a i> l. r i t t i n o f t h e c o ne d i n g e n e s s s p e c i b l e. m o a n d i r g e n p b a g
\end{quote}

We observed the pattern that when the token length is small, e.g., 1, the candidate LLMs may just generate single characters or meaningless symbols but not words, which would lead to cascade effects of generating low-quality completion results. To mitigate the problem, one simple-yet-effect solution is to add the buffer tokens, i.e. if we want one more token in each iteration, we ask the candidate LLMs to generation \textit{(1+k)} tokens and select the first word to append for that iteration, where the buffer number \textit{k} can be set as different integer values. 
We test different \textit{k} values from 1 to 5 in the sQuAD dataset and we find the value \textit{k=3} suits a good balance between better generation quality and longer generation time consumption. One example is listed below (sQuAD dataset, classic LLMs set, token length for 1, 3 buffer tokens, means that we generate 4 tokens in each token and append only the first one):

\begin{quote}
   The success of its football team made Notre Dame a household name . The success of Note Dame reflected rising status of Irish Americans and Catholics in the 1920s . Despite being based in New York, Notre Dame was a highly religious institution, one that recruited and groomed future U.S. presidents while maintaining `` a nonsectarian atmosphere.. During this time, the sport swriting world saw the birth of a magazine called Sport Illustrated, a sports broadcasting conglomerate called Broadcasting-Synd ication Companies, later known as the Ford -Hopkins System, and the National Hockey League .The Irish immigrants who lived through that era became the fathers of such Notre Dame alumni as Notre Dame football coach Lou Holtz, John F. Kennedy, John Lennon, and Jimmy Carter .The Irish immigrants were not alone in celebrating their school. The New York Irish-American newspaper published a special issue in April, 1927, commemorating Irish -Americans in sport. The feature on Notre Dame football
\end{quote}

\begin{table*}[ht]
    \centering
    \begin{tabular}{p{12mm}|p{27.5mm}|p{12mm}|p{9mm}|p{9mm}|p{9mm}|p{9mm}|p{9mm}|p{9mm}|p{9mm}}
    \toprule
    Datasets & Detection Method & Baseline & TL=1 & TL=2 & TL=3 & TL=4 & TL=5 & Rand. & Sent. \\
    \hline
    \hline
        & Likelihood & 0.7837 & 0.3147 & \textbf{0.2015} & 0.2466 & 0.2664 & 0.2923 & 0.2295 & 0.4492 \\
        & Rank & 0.8068 & 0.3787 & \textbf{0.3625} & 0.4018 & 0.4246 & 0.4520 & 0.3892 & 0.5797\\
    XSum & LogRank & 0.8117 & 0.3877 & \textbf{0.2827} & 0.3307 & 0.3572 & 0.3792 & 0.3165 & 0.5191 \\
        & Entropy & \textbf{0.5300} & 0.6983 & 0.8068 & 0.8209 & 0.8269 & 0.8106 & 0.8177 & 0.7435 \\
        & GPT-2 Detector & 0.9416 & 0.7248 & 0.6261 & 0.6130 & \textbf{0.6048} & 0.6402 & 0.6320 & 0.7025 \\
        & RADAR & 0.8904 & 0.9647 & 0.9603 & 0.9347 & 0.9426 & 0.9352 & 0.9438 & 0.8963 \\
        & Fast-DetectGPT & 0.9845 & 0.4573 & \textbf{0.4431} & 0.5653 & 0.6288 & 0.6406 & 0.5245 & 0.8062 \\
    \hline
    \hline
        & Likelihood & 0.7573 & \textbf{0.2602} & 0.2626 & 0.3237 & 0.3757 & 0.3783 & 0.2986 & 0.5493 \\
        & Rank & 0.7836 & \textbf{0.3684} & 0.4212 & 0.4586 & 0.5216 & 0.4806 & 0.4474 & 0.6224 \\
    sQuAD & LogRank & 0.8090 & \textbf{0.3657} & 0.3877 & 0.4535 & 0.4955 & 0.4961 & 0.4232 & 0.6357 \\
        & Entropy & \textbf{0.5617} & 0.7721 & 0.8149 & 0.8049 & 0.7801 & 0.7724 & 0.8025 & 0.7152 \\
        & GPT-2 Detector & 0.9335 & 0.6050 & \textbf{0.5649} & 0.5767 & 0.6074 & 0.5981 & 0.5806 & 0.7107 \\
        & RADAR & 0.8263 & 0.9336 & 0.9124 & 0.8730 & 0.8912 & 0.8831 & 0.8858 & 0.8067 \\
        & Fast-DetectGPT & 0.9937 & \textbf{0.5068} & 0.6035 & 0.7002 & 0.7565 & 0.7541 & 0.6810 & 0.9062\\
    \hline
    \hline
        & Likelihood & 0.8905 & 0.7131 & 0.6780 & 0.6965 & 0.7027 & 0.7075 & \textbf{0.6727} & 0.7866 \\
        & Rank & 0.8186 & 0.6542 & \textbf{0.6458} & 0.6702 & 0.6671 & 0.6725 & 0.6527 & 0.7145 \\
    Writing & LogRank & 0.9158 & 0.7728 & \textbf{0.7490} & 0.7650 & 0.7683 & 0.7705 & 0.7426 & 0.8305\\
        & Entropy & \textbf{0.3752} & 0.4357 & 0.5449 & 0.5857 & 0.5969 & 0.5934 & 0.5981 & 0.5135 \\
        & GPT-2 Detector & 0.9365 & 0.8100 & 0.7534 & 0.7234 & \textbf{0.7094} & 0.7400 & 0.7352 & 0.7437 \\
        & RADAR & 0.7855 & 0.9226 & 0.8597 & 0.8401 & 0.8161 & 0.8307 & 0.8562 & 0.7761 \\
        & Fast-DetectGPT & 0.9977 & \textbf{0.7817} & 0.8718 & 0.9253 & 0.9321 & 0.9364 & 0.8996 & 0.9581 \\
        \hline
    \bottomrule
    \end{tabular}
    \caption{All AI content detection metric AUROC scores for the XSum, sQuAD, and Writing datasets, reported in various \textsc{ToBlend} generation settings. Baseline scores come from the average score for each dataset in Table~\ref{tab-replication}. Here, the candidate models are GPT-2, OPT, GPT-Neo, and GPT-J. Here we ask the model to generate the exact amount of tokens we need in each completion iteration, without extra buffer tokens.}
    \label{tab-old-benchmark-all-metrics}
\end{table*}

\begin{table*}[ht]
    \centering
    \begin{tabular}{p{12mm}|p{27.5mm}|p{12mm}|p{9mm}|p{9mm}|p{9mm}|p{9mm}|p{9mm}|p{9mm}|p{9mm}}
    \toprule
    Datasets & Detection Method & Baseline & TL=1 & TL=2 & TL=3 & TL=4 & TL=5 & Rand. & Sent. \\
    \hline
    \hline
        & Likelihood & 0.7837 & 0.0977 & \textbf{0.0485} & 0.0809 & 0.0970 & 0.1331 & 0.0640 & 0.3023 \\
        & Rank & 0.8068 & \textbf{0.1980} & 0.1998 & 0.2416 & 0.2720 & 0.2955 & 0.2138 & 0.4642\\
    XSum & LogRank & 0.8117 & 0.1334 & \textbf{0.0777} & 0.1217 & 0.1417 & 0.1835 & 0.0973 & 0.3389 \\
        & Entropy & 0.5300 & 0.8780 & 0.9339 & 0.9047 & 0.8842 & 0.8628 & 0.9237 & 0.7495 \\
        & GPT-2 Detector & 0.9416 & 0.6391 & 0.4632 & \textbf{0.4591} & 0.4717 & 0.4839 & 0.4804 & 0.5520 \\
        & RADAR & 0.8904 & 0.9826 & 0.9507 & 0.9602 & 0.9601 & 0.9545 & 0.9555 & 0.9075 \\
        & Fast-DetectGPT & 0.9845 & 0.2191 & \textbf{0.2139} & 0.2890 & 0.2985 & 0.3400 & 0.2576 & 0.5382 \\
    \hline
    \hline
        & Likelihood & 0.7573 & \textbf{0.0638} & 0.0960 & 0.1618 & 0.1965 & 0.2195 & 0.1465 & 0.4800 \\
        & Rank & 0.7836 & \textbf{0.1883} & 0.2387 & 0.3009 & 0.3168 & 0.3333 & 0.2722 & 0.5286 \\
    sQuAD & LogRank & 0.8090 & \textbf{0.0985} & 0.1507 & 0.2365 & 0.2699 & 0.2954 & 0.2168 & 0.5360 \\
        & Entropy & 0.5617 & 0.9095 & 0.8785 & 0.8249 & 0.8122 & 0.7891 & 0.8345 & 0.6290 \\
        & GPT-2 Detector & 0.9335 & 0.4847 & \textbf{0.3507} & 0.3686 & 0.3784 & 0.3925 & 0.3688 & 0.5368 \\
        & RADAR & 0.8263 & 0.9710 & 0.9111 & 0.9094 & 0.9092 & 0.9075 & 0.9032 & 0.8380 \\
        & Fast-DetectGPT & 0.9937 & \textbf{0.2051} & 0.2742 & 0.3656 & 0.4458 & 0.4531 & 0.3531 & 0.7046\\
    \hline
    \hline
        & Likelihood & 0.8905 & \textbf{0.4223} & 0.4828 & 0.5295 & 0.5801 & 0.5781 & 0.4975 & 0.7126 \\
        & Rank & 0.8186 & 0.5154 & \textbf{0.5327} & 0.5695 & 0.5972 & 0.5900 & 0.5390 & 0.6641 \\
    Writing & LogRank & 0.9158 & 0.4933 & \textbf{0.5501} & 0.5965 & 0.6414 & 0.6382 & 0.5658 & 0.7487\\
        & Entropy & 0.3752 & 0.6338 & 0.6515 & 0.6400 & 0.5947 & 0.6051 & 0.6388 & 0.5102 \\
        & GPT-2 Detector & 0.9365 & 0.6816 & 0.5686 & \textbf{0.5684} & 0.6030 & 0.5925 & 0.5772 & 0.6230 \\
        & RADAR & 0.7855 & 0.9312 & 0.8547 & 0.8402 & 0.8301 & 0.8205 & 0.8566 & 0.8056 \\
        & Fast-DetectGPT & 0.9977 & \textbf{0.5734} & 0.7134 & 0.7727 & 0.7916 & 0.7939 & 0.7170 & 0.8846 \\
        \hline
    \bottomrule
    \end{tabular}
    \caption{All AI content detection metric AUROC scores for the XSum, sQuAD, and Writing datasets, reported in various \textsc{ToBlend} generation settings. The difference is that the candidate models used here are Llama2, Phi-2, Mistral, and Gemma. Here we ask the model to generate the exact amount of tokens we need in each completion iteration, without extra buffer tokens.}
    \label{tab-old-benchmark-better-all-metrics}
\end{table*}

\begin{table*}[ht]
    \centering
    \begin{tabular}{p{12mm}|p{27.5mm}|p{12mm}|p{9mm}|p{9mm}|p{9mm}|p{9mm}|p{9mm}|p{9mm}|p{9mm}}
    \toprule
    Datasets & Candidate LLMs & Baseline & TL=1 & TL=2 & TL=3 & TL=4 & TL=5 & Rand. & Sent. \\
    \hline
    \hline
    XSum    & Classic LLMs & \textbf{4.3}/\textbf{4.8} & 3.9/3.4 & 3.4/3.6 & 4.0/4.2 & 2.9/3.3 & 3.7/4.2 & 3.5/3.4 & 3.4/4.0 \\
        & Advanced LLMs & \textbf{4.3}/\textbf{4.8} & 2.0/1.5 & 3.9/3.4 & 3.3/3.1 & 3.7/3.6 & 4.2/3.7 & 3.6/3.7 & 3.9/4.3\\
    \hline
    \hline
    sQuAD    & Classic LLMs & 5.2/\textbf{5.7} & 4.0/3.8 & 5.1/5.3 & 4.1/4.9 & 4.1/4.6 & 4.3/4.9 & \textbf{5.4}/5.3 & 4.7/5.6 \\
        & Advanced LLMs & \textbf{5.2}/\textbf{5.7} & 3.1/3.5 & 3.5/3.9 & 4.3/4.9 & 4.7/5.0 & 3.7/4.3 & 4.8/4.8 & 4.7/5.6\\
    \hline
    \hline
    Writing    & Classic LLMs & \textbf{4.3}/\textbf{4.0} & 2.8/3.2 & 2.4/2.9 & 2.9/2.9 & 3.4/3.9 & 2.8/2.7 & 2.8/2.5 & 2.8/3.4 \\
        & Advanced LLMs & \textbf{4.3}/\textbf{4.0} & 3.4/3.4 & 4.2/\textbf{4.0} & 3.4/3.1 & 3.3/3.9 & 2.2/2.5 & 3.3/2.8 & 3.2/3.2 \\
        \hline
    \bottomrule
    \end{tabular}
    \caption{The average coherence/fluency scores for sampled XSum, sQuAD, and Writing sub-datasets, annotated by three hired human experts and recorded in various \textsc{ToBlend} generation settings. Here we ask the model to generate the exact amount of tokens we need in each completion iteration, without extra buffer tokens.}
    \label{tab-old-human-evaluation}
\end{table*}

\begin{table*}[ht]
    \centering
    \begin{tabular}{p{12mm}|p{27.5mm}|p{12mm}|p{9mm}|p{9mm}|p{9mm}|p{9mm}|p{9mm}|p{9mm}|p{9mm}}
    \toprule
    Datasets & Detection Method & Baseline & TL=1 & TL=2 & TL=3 & TL=4 & TL=5 & Rand. & Sent. \\
    \hline
        \hline
        & DetectGPT & 0.9079 & 0.8112 & 0.8496 & 0.8076 & 0.8532 & \textbf{0.7863} & 0.8140 & 0.8452 \\
    XSum    & LLR & 0.8451 & 0.6720 & 0.6528 & \textbf{0.6152} & 0.6348 & 0.6500 & 0.6216 & 0.7068 \\
        & NPR & 0.9067 & 0.8564 & 0.8564 & 0.8456 & 0.8512 & 0.8136 & \textbf{0.8116} & 0.8484 \\
    \hline
    \hline
        & DetectGPT & 0.8625 & 0.7344 & 0.8090 & 0.8133 & 0.8033 & \textbf{0.6844} & 0.8356 & 0.6989 \\
    sQuAD    & LLR & 0.8856 & \textbf{0.6722} & 0.7467 & 0.7544 & 0.7200 & 0.7644 & 0.7756 & 0.7633 \\
        & NPR & 0.8897 & \textbf{0.6078} & 0.8489 & 0.8389 & 0.8211 & 0.7478 & 0.8633 & 0.7644 \\
    \hline
    \hline
        & DetectGPT & 0.9202 & \textbf{0.7864} & 0.8472 & 0.8640 & 0.8788 & 0.8760 & 0.8896 & 0.8968 \\
    Writing    & LLR & 0.9734 & \textbf{0.9152} & 0.9372 & 0.9204 & 0.9428 & 0.9516 & 0.9180 & 0.9476 \\
        & NPR & 0.9463 & \textbf{0.8480} & 0.9324 & 0.9128 & 0.9278 & 0.9280 & 0.9424 & 0.9240 \\
        \hline
    \bottomrule
    \end{tabular}
    \caption{Result of the classic LLMs set---AUROC scores for the XSum, sQuAD, and Writing datasets by various \textsc{ToBlend} generation settings. Specifically for the detection method of DetectGPT~\cite{pmlr-v202-mitchell23a} and DetectLLM (LLR and NLP)~\cite{su2023detectllm} scores. The candidate models used here are GPT-2, OPT, GPT-Neo, and GPT-J.}
    \label{tab-benchmark-three_more-all-metrics}
\end{table*}

\begin{table*}[ht]
    \centering
    \begin{tabular}{p{12mm}|p{27.5mm}|p{12mm}|p{9mm}|p{9mm}|p{9mm}|p{9mm}|p{9mm}|p{9mm}|p{9mm}}
    \toprule
    Datasets & Detection Method & Baseline & TL=1 & TL=2 & TL=3 & TL=4 & TL=5 & Rand. & Sent. \\
    \hline
    \hline
        & DetectGPT & 0.9079 & \textbf{0.4632} & 0.5672 & 0.6196 & 0.6616 & 0.5644 & 0.5740 & 0.6320 \\
    XSum    & LLR & 0.8451 & 0.4860 & \textbf{0.4172} & 0.4992 & 0.4636 & 0.4484 & 0.4700 & 0.4988 \\
        & NPR & 0.9067 & \textbf{0.5436} & 0.5688 & 0.6648 & 0.6640 & 0.6264 & 0.6104 & 0.6400 \\
    \hline
    \hline
        & DetectGPT & 0.8625 & \textbf{0.4578} & 0.5267 & 0.6033 & 0.6600 & 0.4989 & 0.5233 & 0.6511 \\
    sQuAD    & LLR & 0.8856 & 0.6078 & 0.6433 & 0.6111 & 0.6644 & \textbf{0.5867} & 0.5989 & 0.6433 \\
        & NPR & 0.8897 & \textbf{0.5033} & 0.5778 & 0.6333 & 0.7156 & 0.5444 & 0.5378 & 0.6122 \\
    \hline
    \hline
        & DetectGPT & 0.9202 & \textbf{0.4844} & 0.6784 & 0.7648 & 0.7044 & 0.7284 & 0.7504 & 0.8120 \\
    Writing    & LLR & 0.9734 & 0.8592 & 0.8648 & 0.8696 & 0.8736 & \textbf{0.8192} & 0.8504 & 0.8928 \\
        & NPR & 0.9463 & \textbf{0.5900} & 0.7472 & 0.8088 & 0.7760 & 0.7916 & 0.7840 & 0.8560 \\
        \hline
    \bottomrule
    \end{tabular}
    \caption{Result of the advanced LLMs set---AUROC scores for the XSum, sQuAD, and Writing datasets by various \textsc{ToBlend} generation settings. Specifically for the detection method of DetectGPT~\cite{pmlr-v202-mitchell23a} and DetectLLM (LLR and NLP)~\cite{su2023detectllm} scores. The difference is that the candidate models used here are Llama2, Phi-2, Mistral, and Gemma.}
    \label{tab-benchmark-better-three_more-all-metrics}
\end{table*}

\begin{table*}[ht]
    \centering
    \begin{tabular}{|p{150mm}|}
    \toprule
    Instructions for each instance, you should: 
    \newline \newline
    1. Score from 1 to 7 (1 means poor, 7 means excellent) on the coherence score and fluency score for the column of 'a', 'b', 'c', 'd', 'e', 'f', 'g', 'h'.
    \newline
    * Coherence: The contextual information of the given short text (around 200 words) should logically make sense, i.e., maintaining topic consistency and logical sequence. 
    * Fluency: The text should read naturally, mimicking the style and syntax of human natural language.
    \newline
    2. From the column of 'a', 'b', 'c', 'd', 'e', 'f', 'g', 'h', select the best one that you think has the highest probability that writes by a real human and put the number in the column of 'best' (Normally, the best one should be the one that possesses the highest coherence and fluency scores you annotated in the previous step).
    \newline \newline
    Do note that:
    \newline
    1. We have 6 files and 5 instances each. Completing all labeling, for one instance, should take less than 6 mins. But feel free to take more time if necessary.
    \newline
    2. All texts selected are only a slice of 200 tokens of their original source, so please do not consider the potential incomplete sentence at the end of the text as one of your scoring criteria.
    \newline \newline
    Examples:
    \newline
    1. [As Muslim institutions of higher learning , the madrasa had the legal designation of waqf . In central and eastern Islamic lands , the view that the madrasa , as a religious trust for pious educational endeavors , is the institutional and social prec ursory for the mosque ( Dar ul -Kh air - ) is prevalent ( Ibd ah 198 5 ; N. A hmad 198 7 ) . The madrasa served as a place of religious instruction and a center for m ak tab ( schools ) . It provided lodging , board , and medical care for the instructors and m ustaf a ( students ) . The madrasa was both a place of instruction for religious m at terial and a living environment . The madrasa therefore served two purposes : the disse mination and perpet uation of the teachings of the Islamic faith and the pro p aga tion of Islamic culture and heritage . The mad rasa was the first step in the path to higher education] [Coherence: 4, Fluency: 5]
    \newline 
    2. [Boston has a continental climate with some maritime influence , and using the −3 C ( 27 F ) coldest month ( January ) isotherm , the city lies within USDA hardiness zone 5 b , with an average annual minimum temperature of around −2 .2 C ( 27 F ) . Bostons climate is compar atively warm for the latitude due to its location within the Northeastern United States . Boston is often identified as a coastal city , and experiences regular and strong effects of maritime climate . However , since Boston is far from the most eastern coastline of the state, its climate has little maritime influence . The effects of the ocean can be seen in the average rainfall rate ( around 4 3 inches or 1 09 centim et ers of snow per year ) , a rain shadow that prevents heavy rainfall from accumulating in the summer and a generally war mer average annual temperate , but the city is rarely affected by extreme cold ,] [Coherence: 6, Fluency: 7]
    \newline 
    3. [South Korea : The event was held in Seoul , which hosted the 1988 Summer Olympics , on April 27 . Intended torchbearers Choi Seung-kook and Park Won-sun boycotted the games . To demonstrate their disapproval, many South Koreans wore black rib ons to mourn the massacre. http : //en . w ikinews . org . 30 Nov . 2 0 1 6 . Korea , South - 19 88 Summer Olympics . This website is not to be accredited with any additional information about this event . The images were taken and used in this website by the users . http : // www . z iman . co . kr . h o d oj 69 0 . 46 . 01 0 . ↑ h t t p : //www.al jazeera . co m / program /ins igh t a l/ asia pacif is m / 2 0 1 7/ 0 5 / 01 3  / 8 4 6 5 492 . 27 0162 2 < eos >.] [Coherence: 1, Fluency: 2] \\
    \bottomrule

    \end{tabular}
    \caption{Instruction information we give to our hired human experts to annotate the quality scores for given short texts and select the most human-like generation result from eight candidates for each instance. }
    \label{tab:instruction_human_experts}
\end{table*}

\section{More Detection Models}
\label{sec:appendix-more-detection-models}
As mentioned above, we report three more AI-generated detection approaches from DetectGPT~\cite{pmlr-v202-mitchell23a} and DetectLLM (LLR and NPR scores)~\cite{su2023detectllm} here in Table~\ref{tab-benchmark-three_more-all-metrics} and Table~\ref{tab-benchmark-better-three_more-all-metrics}. 
The LLR score in DetectLLM is more sensitive to machine-generated text than the DetectGPT model.
For machine-generated and human-written texts, the detection performance (e.g., for LogRank) can be affected by small perturbations, the NPR score is designed to be more sensitive towards perturbations on machine-generated texts.
The implementations of DetectGPT and DetectLLM models in our local server are relatively time-consuming, the estimation to run full datasets would be around 400 GPU hours (for A100). So we report the AUROC scores calculated on the sub-sampled 10\% datasets to improve the time efficiency.

As shown in Table~\ref{tab-benchmark-three_more-all-metrics} and Table~\ref{tab-benchmark-better-three_more-all-metrics}, we observe very similar patterns compared to the Fast-DetectGPT in Table~\ref{tab-benchmark-all-metrics} and Table~\ref{tab-benchmark-better-all-metrics}. Still, the difference is that the DetectGPT and DetectLLM (LLR and NPR) approaches are not as good as Fast-DetectGPT in the baseline and generally exhibit larger AUROC score drops when facing our proposed \textsc{ToBlend} generation method.

\section{\textsc{ToBlend} Generation Quality Annotation}
\label{sec:appendix-human_experts}
In addition to the human annotations, we also tried to ask the ChatGPT (version 3.5 and 4) to provide the annotation regarding the ChatGPT's capability to understand the numerical scale and evaluate text quality similar to human performance~\cite{huang2024chatgpt}. We use the exact instructions we give to human experts as the prompt of the ChatGPT input. However, after the manual inspection of the ChatGPT annotations, we find only ChatGPT-4 could understand the task and give out annotations in the format we requested. Still, further inspection showcased that around 20\% of the annotation scores are significantly contrary to human expert annotations. Thus, we do not include annotations from LLMs in this work.

\section{Robustness Test for \textsc{ToBlend} Generation Settings}
\label{sec:appendix-robustness}
As discussed in section~\ref{sec:appendix-generation_length}, adding buffer tokens could help to effectively improve the quality of the \textsc{ToBlend} generation result. However, the \textsc{ToBlend} generation result could still contain a few with meaningless symbols, like the example below (XSum dataset, classic LLMs set, token length for 4, 3 buffer tokens):

\begin{quote}
   Maj Richard Scott, 40, is accused of driving at speeds of up to 95mph (153km/h) in bad weather before the smash on a B-road in Wiltshire, the court heard. Judge Thomas Tull och rejected the defence claims that the Army officer, who also served in Iraq and Afghanistan, may have felt stressed. Mr Scott was remanded in custody for sentencing later this year. < |endoftext| > Posted by Michael McAllister at 11:17 am on 16 April 2012 (US) ¢¬Ä¬î¢¬Ä¬î¢¢ < /s > |¢¬Ä¬î¢¬Ä¬¬Ä¬î | | | |  Maj Richard Scott, 40, is accused of driving at speeds of up to 95mph (153km/ h) in bad weather before the smash on a B-road in Wiltshire, the court heard. Judge Thomas Tull och rejected the defence claims that the Army officer, who also served in Iraq and Afghanistan, may have felt stressed. Mr Scott was remanded in custody for sentencing later this year.
\end{quote}

The \textsc{ToBlend} generation result could also contain some specific expressions that contain LLM-style contents, like \textit{<|endoftext|>}, showcased in the example below (Writing dataset, classic LLMs set, sentence-level blending):

\begin{quote}
    You tell Death that you will never tire of living no matter how long you live. It makes you a bet that you will and thus grants you true power. This goes beyond what is normal for normal beings to believe. You are no normal beings. Death does n't fear you anymore since you are a bet. < |endoftext| >. < /h1 > In all its power you see. < |endoftext| >. < /h1 > All life forms were created by God. < |endoftext| >. < /h1 > It takes a great heart to care for others. < |endoftext| >. < /h1 > < /s >.., < |endoftext| >. < /h2 > The great evil against Death is the belief that it is bad at being. < |endoftext| >. < /h2 > If Death were able to take you away from your loved ones and put you somewhere else for a long period of time, you would be left with guilt. Death could not do this.
\end{quote}

In such case, one feasible solution is to adopt a simple filter function that detects whether the \textsc{ToBlend} generation result contains repetitive single characters, meaningless symbols, or very LLM-style tokens (e.g., \textit{<|endoftext|>}). If so, just re-generate until there is no such content within the \textsc{ToBlend} generation results. We did not adopt such approaches since we want to investigate the \textsc{ToBlend} from a more authentic perspective and showcase its strengths as well as potential failure cases.

Furthermore, we investigate whether the different completion criteria for the text completion would affect the effectiveness of our \textsc{ToBlend} generation approach.
In ~\citet{bao2023fast}, the completion criteria is that the total token length of the completed test should not exceed 200 tokens, including the 30 tokens used in the prompt. 
We attached the robustness test results for our \textsc{ToBlend} generation attack under different completion criteria of not generating more than 100 tokens ended by a period of more than 150 tokens for each instance, as shown in Table~\ref{tab-benchmark-all-metrics-no_length}. The new completion criteria resulted in a more significant AUROC score drop in our \textsc{ToBlend} attack, compared with the \citet{bao2023fast} approaches listed in Table~\ref{tab-old-benchmark-all-metrics}. The new criteria showcased the potential of more effective \textsc{ToBlend} generation settings. 

Still, to make a fair comparison with the human-written texts we used in quality evaluations (we only select the first 200 tokens of the human-written text), we keep AI-generated and human-written text similar in length, in our major experiments. 

\section{Syntactic and Semantic Pattern Analysis during Generation}
\label{sec:appendix-filter-function}
In this section, we will showcase different examples for all three datasets with their coherence and fluency scores reported in Table~\ref{tab-human-evaluation}, under different candidate LLMs sets.
\subsection{Classic LLMs set}
For the XSum dataset, here is the high-quality example generated, token length of random value from 1 to 5, buffer token length of 3, and average coherence/fluency score of 6/6:
\begin{quote}
    The ban was backed by local authorities in Urumqi, state media reported. The move is the latest in a campaign against Islamic clothing and symbols in the mainly Muslim nation of 58 million. Authorities have recently been making similar moves against Islamic fashion shops, and local authorities in the capital, Beijing, have also been seeking to ban the wearing of religious-themed clothes in public places. `` This decision is consistent with the principles of law and administrative regulations and is strictly enforced, `` Xinhua said last month. In the wake of the attack, local officials in Xinjiang have made a series of raids in which they seized religious publications and equipment that have been put up for sale, the state -run China Daily said. Mr Brown said that the latest move by Mr Wisan `` shows the depths of depravity that the Chinese authorities would sink to in order to control every aspect of daily life'and claimed his sister had not been consulted on
\end{quote}

For the XSum dataset, here is the low-quality example generated, token length of 1, buffer token length of 3, and average coherence/fluency score of 3.3/3.3:
\begin{quote}
    She was speaking at an event on Thursday to remember four men who were shot dead by the SAS in her home village of Clonoe in 1992. DUP leader Arlene Foster MLA, says a DUP MP should have been able to visit the soldier who shot her brother. On the day two women were elected to the Irish parliament in the Irish elections last week (both from parties that were once part of a Fianna Fail-led government) it was a political shocker when it came to light that a Fianna Fail minister, John McGuinness, the IRA veteran and a former leader of the Irish Labour party, had lost a local council elections in the same week. But the news that he had lost his seat sent shockwaves through the country. Fianna Fail leader Micheal Martin said he would be stepping down after the party lost the Clonoe byelection. He said his departure was a vote [ Read More ] [
\end{quote}

For the sQuAD dataset, here is the high-quality example generated, token length of 2, buffer token length of 3, and average coherence/fluency score of 5/5.3:
\begin{quote}
    The Bush administration then turned its attention to Iraq, and argued the need to remove Saddam Hussein from power in Iraq had become urgent. Among the stated reasons for the war in Iraq was to eliminate (Iraq 's production of WMD s) WMD s in the region, but also to stop a potentially larger conflict with Iran. This was not a new argument, both the Bush and later the Obama administrations have argued the removal of Saddam from power would end the threat from Iraq. Wary of a new war in Iraq the U.N. Security Council again voted against invading and occupying Iraq. The Bush administration eventually had to admit defeat in the face of strong opposition to its plans for war. And it would be several more years before the international community would get a chance at the real solution to the threat posed by Iraq after the overthrow of Saddam Hussein in 2003 -2006, when the U.S,
\end{quote}

For the sQuAD dataset, here is the low-quality example generated, token length of 2, buffer token length of 3, and average coherence/fluency score of 4.3/4.3:
\begin{quote}
    In practice, not all rule violations will result in a disqualification as the referee may use their own judgement and is not obligated to stop the match. Usually a referee will stop the match if they deem the player has broken the rules to such an extent that their opponent is unable to continue playing, due to the player having a severe injury or has taken a particularly heavy blow from their opponent, and therefore the referee feels the match would have to be stopped if they were to continue. It is the referee ¢¬Ä¬ô s discretion as to whether or not to disqualify a player. If a referee feels the player has broken a rule, they may decide that the player is a no- show, will be given a warning or they may disqualify the opponent and award them another loss. If a referee judges a player has committed a serious injury, they may ask for an additional loss due to the
\end{quote}

For the Writing dataset, here is the high-quality example generated, token length of random value from 1 to 5, buffer token length of 3, and average coherence/fluency score of 5.3/5.7:
\begin{quote}
    You leave a bowl of food out for the dragons. Milk, meat, and potatoes in a huge bowl. More of a bucket, really. Perfect when flying Thank you, what an awesome idea. In the mountains of Tennessee, I have to run into the woods to find food or supplies when I am hungry. It is hard I 'm sorry to hear you lost your job. I 'm also happy to hear you found your happiness Aw ful. Your children are growing and their growth is amazing. That is the best possible gift I could hope for.. to hear your children sing and play music. How nice that is. I wish I had seen this video to your children. Thank you for the kind words. About your chickens, no more worries. The eggs are still good and will probably not give me any health concerns. Oh, you should be very proud
\end{quote}

For the Writing dataset, here is the low-quality example generated, token length of 2, buffer token length of 3, and average coherence/fluency score of 3.3/3.3:
\begin{quote}
    You leave a bowl of food out for the dragons. Milk, meat, and potatoes in a huge bowl. More of a bucket, really. Perfect. One of my personal favorites! They love the stuff! Is there anything that isnt good or awesome about the dragons? Of course. The story! The characters! Everything! They just get everything right, like they just write the best dragons and characters and worlds and stories EVERRRRRRRRR RRRRRRRRRR..... \# 21 The first part in the movie the dragon is chasing after a man. A soldier and some other men in the area. The dragon is chasing him down and getting closer and closer. The soldiers try to get all the people down and get the horses away from the dragon. There is a massive man and a woman. She has a baby in her arms. The baby is the smallest and also the sweetest. This is
\end{quote}

\subsection{Advanced LLMs set}
For the XSum dataset, here is the high-quality example generated, token length of random value from 1 to 5, buffer token length of 3, and average coherence/fluency score of 6/6.3:
\begin{quote}
    She was speaking at an event on Thursday to remember four men who were shot dead by the SAS in her home village of Clonoe in 1992. DUP leader Arlene Foster, and Sinn Fein First Minister Paul Givan, were also in attendance at the village 's Irish language secondary school, St Patrick 's, to hear Irish language poet Nuala NÉ¬= Dhomh nail and read poetry in support of the families of those who died. The four men shot by the Special Forces were republicans Liam Hannaway, Michael Tighe, James Maguire and John Murray. Ms McDonald said: 'This day is really an opportunity for us to remember and to give thanks for their lives. 'Their lives were cut short at the age of 17, 4 and 15, and 30 years today, the community of Clonoe and surrounding areas here want to keep their memory alive 'The pain of their loss has not abated, and the community here today and now
\end{quote}

For the XSum dataset, here is the low-quality example generated, token length of 1, buffer token length of 3, and average coherence/fluency score of 3.7/4:
\begin{quote}
    The ban was backed by local authorities in Urumqi, state media reported. The move is the latest in a campaign against Islamic clothing and symbols in the mainly Muslim area of Xinjiang. In recent years authorities have shut down mosques, closed `` unauthorised `` Qurans and removed women 's headscarves. The ban is also a challenge to the... (Read More) < BR > Muslims in China have been prohibited from wearing Islamic dress, including turbans, veils and long beards, by authorities in the western region of Xinjiang, state media reported Saturday. Read more: http: //www.the independent.co.uk/news /world/asia /muslims -in-ch ina-urged -to-g row-more- pork-to- pros e-islam ic-ban -102524 89.h cms http: //www.the independe nt.co.uk/news /world/asia /muslims -in-ch ina-urged -to- g ro e-more- -102 524 -89.h cms < / BR > < BR > The ban was backed by local authorities in Urumqi, the state -run Xinhua news agency reported. The
\end{quote}

For the sQuAD dataset, here is the high-quality example generated, token length of random value from 1 to 5, buffer token length of 3, and average coherence/fluency score of 5.7/5.7:
\begin{quote}
    An adolescent 's environment plays a huge role in their identity development. While most adolescent studies are conducted on white, middle class children, studies show that the influence of peers and media may have a stronger effect on identity formation among African American youth. Researchers found that when adolescents spend more time with peers and less time with their families, they are more likely to develop their identities around their friends ' values and social roles rather than their own parents ' values and roles Adolescent Identity Development Essay. It has also been found that adolescents exposed to negative peer influences are at a greater risk of engaging in risky behaviors. In her paper `` The Relationship Between Adolescent Peer Association, Peer Pressure, and Adolescent Identity Development, `` Lizzie C. M. Lewis-Brown argues that the pressure exerted by peer groups prompts many teens into exploring their individuality, but this can also lead to adverse effects, such as rebellion, depression
\end{quote}

For the sQuAD dataset, here is the low-quality example generated, token length of 3, buffer token length of 3, and average coherence/fluency score of 3.7/4.3:
\begin{quote}
    In 1954, major Serbian and Croatian writers, linguists and literary critics, backed by Matica srpska and Matica hrvatska signed the Novi Sad Agreement, which in its preamble stated that it would protect and further the Serbian culture in all its manifestations, language, arts, religious faith, literature, historical research and so forth, and that it would equally represent both the Serbian and the Croatian cultures. At the same time, the agreement stated that the Croatian language would be ¢¬Ä¬ú the equal national language of Croatian Serbs in all spheres of national and local life and state institutions ¢¬Ä¬ù. A large number of Croatian, Bosnia k and Muslim leaders rejected that agreement. The first large party to support it, in terms of members, was the Socialist Autonomous Region of Vojvodina. For the first time, this meant that the Serbs in the HSS had two major allies in Croatia and Yugoslavia. This created a political division within
\end{quote}

For the Writing dataset, here is the high-quality example generated, token length of 3, buffer token length of 3, and average coherence/fluency score of 5/5:
\begin{quote}
    You leave a bowl of food out for the dragons. Milk, meat, and potatoes in a huge bowl. More of a bucket, really. Perfect sized. Its the only food I can think of and get the hang of the recipe. And no matter how much I fill it, by tomorrow its almost empty. There is at least half a ton of bones in the huge cave, mostly dragon bones. They will keep for quite some time I imagine. Just wonder how its done. If it involves a large fire. To keep warm, right? Yes... I 've never had pork chops. Its too sweet really. Never tried it. I think I had beef when I was very little. No milk... I hate milk. It 's hard on my stomach. No potatoes... No. Too salty. I think I got potatoes from a bowl of something I had.
\end{quote}

For the Writing dataset, here is the low-quality example generated, sentence-level blending, and average coherence/fluency score of 4.3/4.3:
\begin{quote}
    You leave a bowl of food out for the dragons. Milk, meat, and potatoes in a huge bowl. More of a bucket, really. Perfect for the 300+ kilograms of muscle and bone that your three black-leathered, bright-red scaly babies like to eat. A little milk is good for their hair and coats ; but there is always too much to drink if not supplied in a bucket, as well as, of course, the, potatoes, meat, and other assorted food left on the left of your empty. < |endoftext| >. < /p > There are several things I would recommend doing with this site in the first year or so. < /p > I would like to add that these are NOT the things mentioned in the following sections. < /p > 1. The best and most practical advice is that, once you have a good list of things you want to do, you should start your own. That is to say, you should begin to make a list of things you want to do on a regular basis, and keep track of what you accomplish, when, where, and why.
\end{quote}

\section{Llama3.1 QA and Fine-tune Setting}
\label{sec:appendix-llama2}

We prompt the Llama3.1 model with the prompt design below to collect AI-generation text detection classification results:

\begin{quote}
   Please answer whether the given short text is generated by Artificial Intelligence models but not written from real human.
   
   The short text is: \textit{{Generation results.}}
   
   Please answer by Yes, No or Uncertain. And then explain why in shortly in one or two sentences.
\end{quote}

To avoid overfitting the LLM in the fine-tuning process, we created the dataset with 37 annotated high-quality AI-generated texts with 37 human-written texts with the same source distribution (28 from sQuAD, 7 from XSum, and 2 from Writing). We adopt the template below to utilize the human-written and high-quality AI-generated texts:

\begin{quote}
    \#\#\# Question: Please answer whether the given short text is generated by Artificial Intelligence models but not written from real human.
    
    The short text is:\textit{{instance text}}.
    
    Please answer by Yes, No or Uncertain. And then explain why shortly in one or two sentences.
    
    \#\#\# Answer:\textit{{instance label}}. Yes means the short text is more likely to be generated by AI models but not written by real human. No means the contrary.
\end{quote}

As for the hyper-parameter setting for the fine-tuning process, we fine-tuned the Llama3.1 model at one GPU server containing eight A100 80GB GPUs. We set the lora\_alpha value as 16 and the lora\_dropout as 0.1 for LoRA; we set the optimizer as pages\_adamw\_32bit, learning rate as 0.0002, and weight decay as 0.001 for 5 epochs.

\end{document}